\documentclass{article}

\usepackage[final, nonatbib]{nips_2018}

\usepackage[utf8]{inputenc} 
\usepackage[T1]{fontenc}    
\usepackage{hyperref}       
\usepackage{url}            
\usepackage{booktabs}       
\usepackage{amsfonts}       
\usepackage{nicefrac}       
\usepackage{microtype}      
\usepackage{graphicx}
\usepackage{amsmath}
\usepackage[ruled,vlined]{algorithm2e}
\usepackage[symbol]{footmisc}

\title{Effective Network Compression Using Simulation-Guided Iterative Pruning}

\newcommand*\samethanks[1][\value{footnote}]{\footnotemark[#1]}

\author{
  Dae-Woong Jeong\\
  Information and Electronics Research Institute, KAIST\\
  \texttt{daewoong.jeong@kaist.ac.kr}
  \And
  Jaehun Kim\thanks{The authors contributed equally to this work.} \\
  School of Computing, KAIST \\
  \texttt{jaehunkim@kaist.ac.kr}
  \And
  Youngseok Kim\samethanks \\
  NOTA Incorporated \\
  \texttt{tonykim@nota.ai}
  \And
  Tae-Ho Kim \\
  KAIST Institute for AI, KAIST \\
  \texttt{ktho22@kaist.ac.kr}
  \And
  Myungsu Chae\thanks{Corresponding author.} \\
  NOTA Incorporated \\
  \texttt{mschae89@nota.ai}
}

\begin{document}

\maketitle

\begin{abstract}
 Existing high-performance deep learning models require very intensive computing. For this reason, it is difficult to embed a deep learning model into a system with limited resources. In this paper, we propose the novel idea of the network compression as a method to solve this limitation. The principle of this idea is to make iterative pruning more effective and sophisticated by simulating the reduced	network. A simple experiment was conducted to evaluate the method; the results showed that the proposed method achieved higher performance than existing methods at the same pruning level.
\end{abstract}

\section{Introduction}

Advances in deep neural networks have heavily contributed to the recent popularity of AI. Most algorithms that exhibit state-of-the-art performance in various fields \cite{simonyan2014very, chae2018end} are based on deep neural networks. However, it is difficult to implement deep neural networks without utilizing high-end computing because of their complex and massive network structure. To supply the computing power, most of the existing products based on deep neural networks are processed by high-end servers, which brings about three critical limitations on latency, network cost, and privacy. Therefore, it is necessary to implement deep neural networks on independent clients rather than on servers. Network compression technology is becoming very important as a tool to achieve this.

Various studies on network compression have been extensively performed \cite{he2017channel, lee2018snip, hu2016network, li2016pruning, guo2016dynamic, hubara2017quantized, wu2016quantized, zhou2017incremental, ullrich2017soft}. Additionally, among existing network compression methods, iterative pruning is one of the most renowned methods, as it has been proven to be effective in several previous studies \cite{han2015deep, han2015learning, iandola2016squeezenet}, and is considered to be a state-of-the-art technique \cite{han2015deep}. In the iterative pruning process, the importance of weights in the original network are first evaluated; then, the weights with low importance are removed prior to retraining of the remaining weights for fine-tuning. This pruning process is iteratively performed until the stopping condition is satisfied. However, in this process, the importance of weights is solely determined as based on the original network; thus, it is possible that the pruned weight may be appropriate for implementation in the retrained network. Therefore, we propose a more effective and sophisticated weight pruning method that is based on simulation of a reduced network.

\section{Simulation-Guided Iterative Pruning}

In this section, the proposed method of simulation-guided iterative pruning for effective network compression is introduced. Algorithm \ref{alg:1} shows the detailed algorithm of the proposed method. The main distinction from the baseline model \cite{han2015deep} is that the proposed method utilizes the temporarily reduced network in the simulation.

In the first step of the simulation, the importance of each weight in the original network is calculated, and the original network itself is separately stored in the memory. Then, the weights in the original network with importance below a certain threshold are temporarily set to zero to create a temporarily reduced network. Here, a predetermined percentile threshold, rather than the threshold in the baseline method, is used to ensure consistency throughout the iterative simulation process. The gradients for each weight of the temporarily reduced network, including zero weights, are then calculated by using a batch of training data. These gradients are applied to the stored original network rather than the reduced network. Then, this simulation process begins again from the first step, incorporating the changes made to the original network, and repeats until the network is sufficiently simulated.



\begin{algorithm}[t] \label{alg:1}
\DontPrintSemicolon
\SetKwInOut{Input}{Input}\SetKwInOut{Output}{Output}
\Input{Pre-trained neural network model $M$, Pruning steps $n$, Pruning ratio $r$, Simulation steps $s$}
\Output{Reduced network $R$}
Initialize zero position matrix $Z$\;
$M_{1}$ = $M$\;
\While{$pruning$ $step<n$}{
\While{$simulation$ $step<s$}{
  Generate temporarily reduced network $T$ by pruning in $M_{pstep}$ using ratio $r$\;
  Calculate gradient $g$ using $T$\;
  \For{Parameters in $M_{pstep}$} {
   \If{Parameter position is not in $Z$} {
   Update $Parameter$ in model $M_{pstep}$ using $g$\;
   }
  }
}
Add parameter positions to $Z$ using ratio $r$\;
$M_{pstep + 1} = M_{pstep}$\;
}
$R = M_{n}$\;
\caption{Simulation-guided iterative pruning}
\end{algorithm}

After this simulation process, the importance of the weights are computed, and weights below the threshold are removed as described above for the iterative pruning method. Then, the pruned weights are permanently fixed, and the entire process is repeated with a higher threshold and no retraining process.


\begin{figure}[h] 
  \includegraphics[scale=0.85]{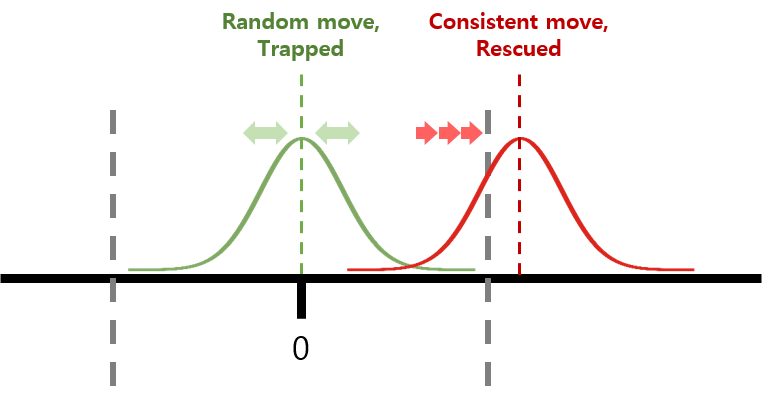}
  \centering
  \caption{Conceptual diagram showing the simulation-guided iterative pruning process.}
  \label{fig:1}
\end{figure}

The proposed approach can supplement the limitations of the baseline model. Figure \ref{fig:1} conceptualizes the iterative simulation process; the horizontal axis corresponds to weight values, and the grey dashed lines indicate the pruning threshold. The green dashed line indicates the ideal value of a truly insignificant weight that is positioned near zero; the red dashed line indicates the ideal value of a weight that is considered to be significant. Since the learning process of the network implements a stochastically chosen batch of data, the ideal value of a significant weight is also stochastically distributed near the absolute ideal value, as is shown in the figure. In this case, even if the absolute ideal value of the significant weight is larger than the threshold, the value of the weight could be undesirably categorized as within the cut-off range. Pruning of this particular type of weight results in unnecessary loss of information. Therefore, it is important to distinguish between the truly insignificant weights and significant but miscategorized weights. Here, when the weight value is set to zero during the simulation, the gradient of the insignificant weight has a randomized direction because the absolute ideal value is sufficiently close to zero. On the other hand, the direction of the gradient of the significant weight is unchanged until the weight is rescued through the iterative simulation. Moreover, unlike the baseline method, this screening process relies on simulation of a pruned network, and thus enables more sophisticated pruning.


\section{Experiment}

\subsection{Experimental Setup}

For the experiment, we respectively applied the proposed and baseline algorithms to MNIST data \cite{lecun1998gradient}. The network architecture used for each algorithm was a fully-connected LeNet \cite{lecun1998gradient}. The architecture comprises three hidden layers, with each layer respectively consisting of 784, 300, and 100 nodes in sequence. The final output layer consists of 10 nodes.

In the experiment, the batch size was set as 50, and the optimizer was the Adam optimizer. The loss term of L2 regularization was 0.0001, and the learning rate was 0.005. Pruning began with a model that was first trained on 10 epochs and a batch size of 50. For each iteration, the percentile threshold for pruning was set as 5\%, and the pruned weight was set to zero.



\subsection{Experimental Result}

\begin{figure}[t]  

  \includegraphics[scale=0.4]{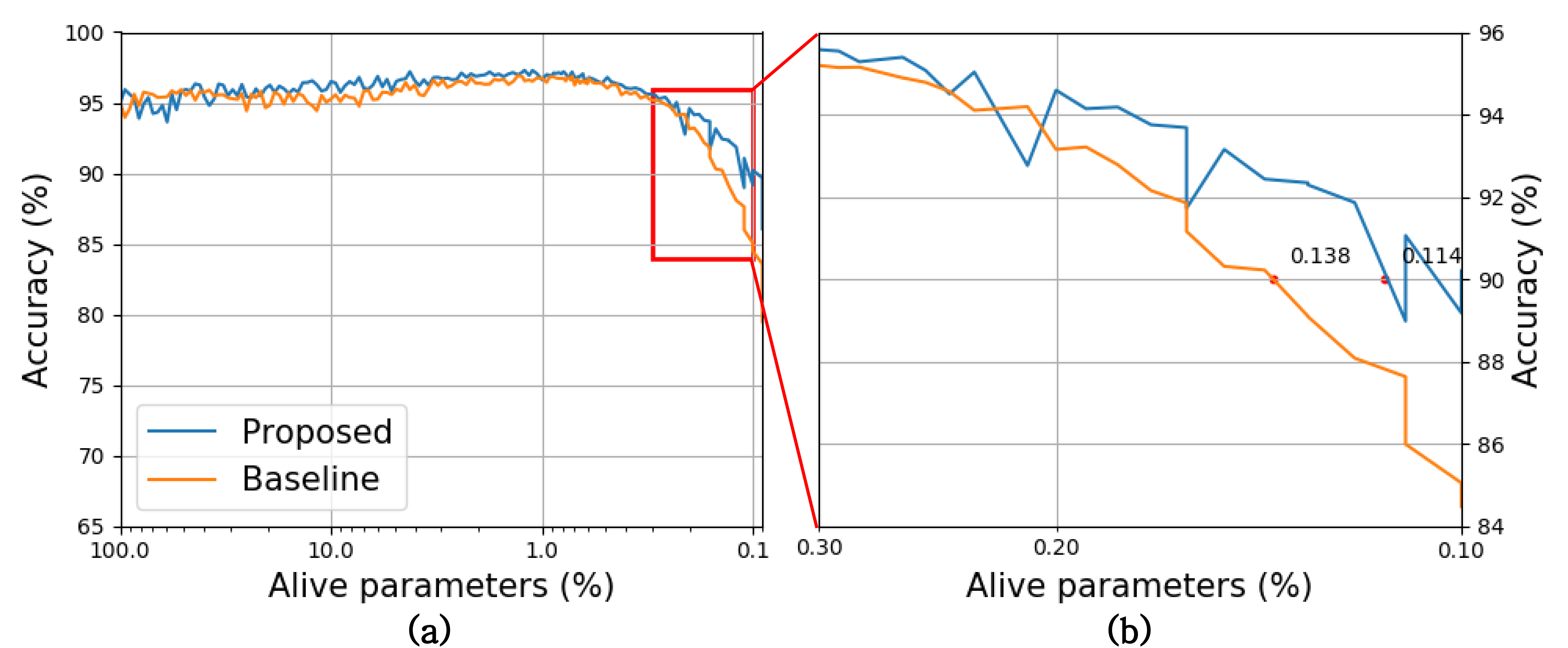}
  \centering
  
  \caption{Experimental results for the proposed and baseline models. (a) Accuracy of the network as a function of the percentile ratio of the remaining weights; (b) Enlargement of the red box in (a).}
  \label{fig:2}
\end{figure}

Figure \ref{fig:2} shows the performances of the proposed and baseline algorithms implemented in this experiment. The experimental results are presented as classification accuracy as a function of the ratio of alive parameters in the pruning process. Figure \ref{fig:2} (a) shows that there is no significant difference in performance, even when the number of parameters is reduced by 1\%. In addition, the performance of the baseline algorithm is drastically degraded as the number of the surviving parameters drops below 1\%, whereas that of the proposed algorithm remains relatively high at the same level.


Figure \ref{fig:2} (b) is an enlargement of the results in the red box of figure \ref{fig:2} (a); the difference between the performances of the two algorithms dramatically increases. The results show that the proposed algorithm performs network compression more effectively than the baseline algorithm at the same performance level. Specifically, the proposed algorithm compressed the network into 0.114\% of its original size while maintaining a classification accuracy of 90\%, whereas the baseline algorithm can only compress the same network into 0.138\% of its original size.


\section{Conclusion}

In this paper, we proposed a novel method to compress deep neural networks. We focused on making the iterative pruning process more effective by simulating a temporarily reduced network. With this method, the reduced network enables collaborative learning of a more suitable structure and optimal weights. The experiment to evaluate the method showed that the proposed algorithm outperforms the baseline algorithm. The proposed method can be used to mount a high-performance deep learning model onto an embedded system with limited resources.


\section*{Acknowledgment}

This work (Grants No. S2520005) was supported by Business for Startup growth and technological development(TIPS Program) funded Korea Ministry of SMEs and Startups in 2018.

\bibliographystyle{unsrt}
\bibliography{reference.bib}

\end{document}